\documentclass[letterpaper, 10 pt, conference]{ieeeconf}

\IEEEoverridecommandlockouts
\usepackage[T1]{fontenc}
\usepackage{amsmath,amssymb,bm}
\usepackage{cite}
\usepackage{graphicx}
\usepackage{booktabs}
\usepackage{balance}
\usepackage{tabularx}
\usepackage{array}
\usepackage{microtype}
\usepackage{url}
\usepackage{color}
\usepackage{stfloats}

\newcolumntype{Y}{>{\raggedright\arraybackslash}X}
\newcommand{\method}{RodForesight}
\newcommand{\bg}{\mathbf{g}}
\newcommand{\ba}{\mathbf{a}}
\newcommand{\bc}{\mathbf{c}}
\newcommand{\bh}{\mathbf{h}}

\newcommand{\bu}{\mathbf{u}}

\newcommand{\by}{\mathbf{y}}
\newcommand{\calL}{\mathcal{L}}

\title{\LARGE \bf
RodForesight: A World Model Enhanced Diffusion Policy for Slender Rod Insertion}

\author{Chuanbo Yu$^{1}$, Mingyu Yue$^{2}$, Yan Lyu$^{3}$, 
Chuhan Song$^{4}$, and Peng Wang$^{5,*}$%
\thanks{$^{1}$Chuanbo Yu is with SWJTU-Leeds Joint School, Southwest Jiaotong University, Chengdu, China. {\tt\small chara@my.swjtu.edu.cn}}%
\thanks{$^{2}$Mingyu Yue is with the School of Mechanical Engineering, University of Leeds, Leeds LS2 9JT, United Kingdom.
{\tt\small hlld8753@leeds.ac.uk}}%
\thanks{$^{3}$Yan Lyu is with Southeast University, Nanjing, China. {\tt\small lvyanly@seu.edu.cn}}%
\thanks{$^{4}$Chuhan Song is with the Department of Computer Science, University of Oxford, Oxford OX1 3QD, United Kingdom. {\tt\small hert7886@ox.ac.uk}}%
\thanks{$^{5}$Peng Wang is with the Centre for Vision, Speech and Signal Processing (CVSSP), University of Surrey, Guildford GU2 7XH, United Kingdom.}%
\thanks{$^{*}$Corresponding author: {\tt\small peng.wang@surrey.ac.uk}}%
}

\begin{document}
\raggedbottom
\maketitle
\thispagestyle{empty}
\pagestyle{empty}

\begin{abstract}
Slender rod insertion arises in precision manufacturing, where millimetre scale diameter and tight clearances demand accurate perception and control. Conventional peg-in-hole methods assume a rigid object whose tip pose is fixed relative to the gripper. This assumption breaks down for a high aspect ratio rod, which can bend during manipulation, making its tip motion dependent on the rod configuration, grasp, material properties, and contact. We present \textit{RodForesight}, a learning framework that factorises the task into two stages: 1) coarse approaching, which uses visual servoing to map diverse initial configurations into a compact near hole hand-off region; and 2) predictive insertion, which performs fine alignment and completes the insertion. It is worth noting that the two stages can be wrapped into an end-to-end design. During insertion, a diffusion policy generates candidate action chunks, while an action conditioned world model predicts their effects on rod-hole alignment. This pre-execution evaluation enables RodForesight to select the best action chunk based on predicted tilt and radial errors before execution. Experiments investigate the performance of different stages and the end-to-end setting, where RodForesight improves the success rate from 88.9\% to 96.7\%, compared to baseline methods such as diffusion policy. 
\end{abstract}

\section{Introduction}

\begin{figure}[t]
    \centering
    \includegraphics[width=\columnwidth]{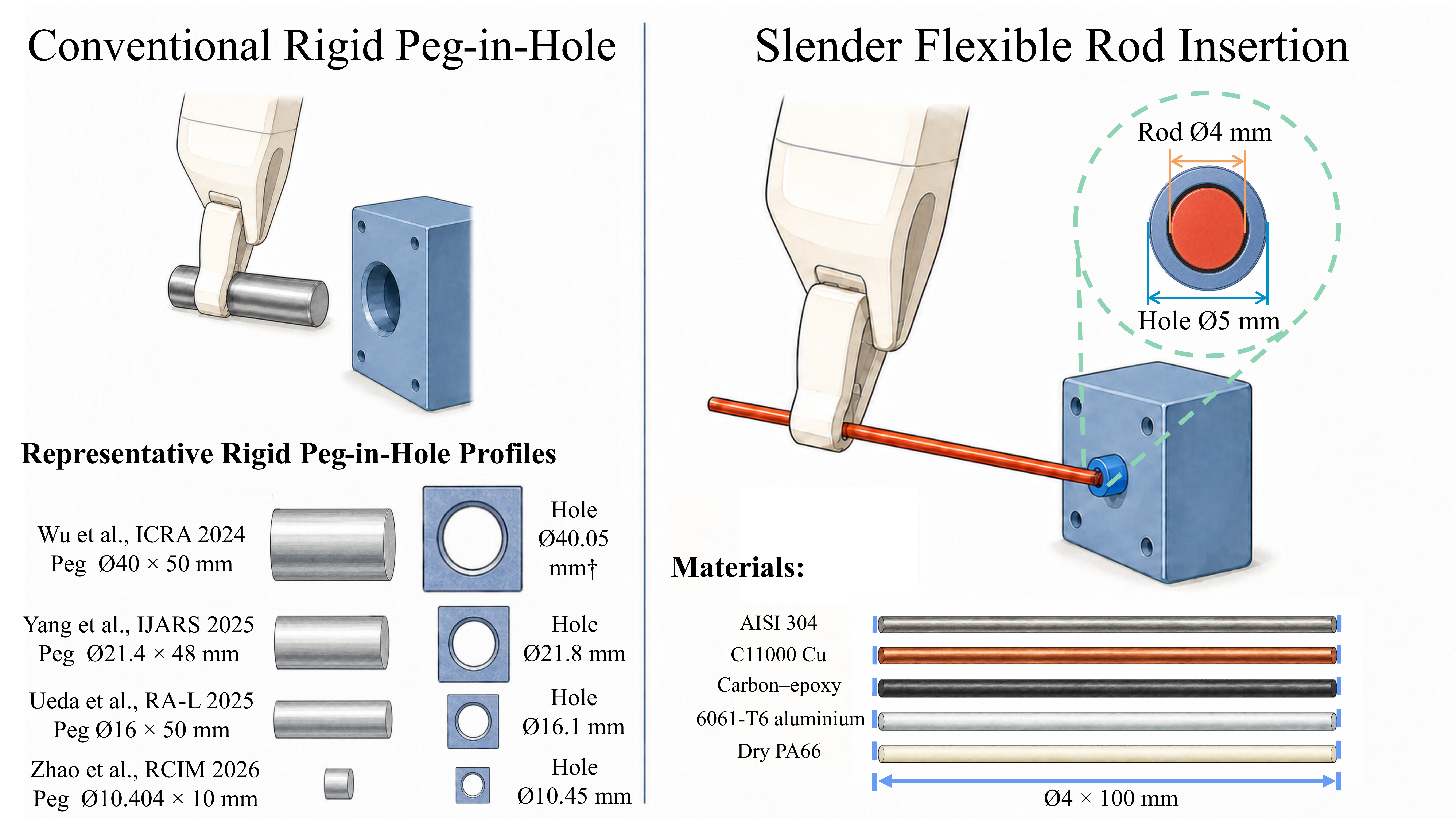}
    \caption{Task and scale contrast. \textit{Left}: rigid-peg dimensions and associated hole sizes compiled from Wu et al.~\cite{wu2024behavior} (inferred volumes), Yang et al.~\cite{yang2025hybrid}, Ueda et al.~\cite{ueda2025robust}, and Zhao et al.~\cite{zhao2026jamming}. \textit{Right}: a 100 mm long with 4 mm diameter slender rod enters a 5 mm diameter, 20 mm deep hole with 0.5~mm radial clearance. The training suite includes AISI~304 stainless steel, C11000 copper, and isotropic carbon-epoxy, while 6061-T6 aluminium and dry PA66 are reserved for generalisation tests. }
    \label{fig:task_contrast}
\end{figure}

The insertion of slender rods requires precise control of the distal position and orientation during approach and contact. Unlike traditional rigid peg-in-hole tasks where robots insert pegs at centimetre level~\cite{activecontact,robustpeg}, the millimetre scale diameter and high aspect ratio (length to diameter) introduce new perception and insertion challenges. Contact sensing and compliance that are widely used in rigid peg-in-hole task completion may break down in slender rods insertion because the reduced volumes make the rods prone to bend and more difficult to detect. Successful insertion of slender rods therefore needs to consider factors ranging from geometry observability, robot action dependent deformation, material response and grasp configuration~\cite{dloinsert,dlodynamics}. Fig.~\ref{fig:task_contrast} shows some representative peg-in-hole task profiles in literature and illustrates the slender rod insertion task in this manuscript.


Learning based approaches have been widely used for insertion due to variations in geometry, material properties, and contact dynamics. Reinforcement learning has been applied to learn insertion policies through interaction, while flexibility-conditioned methods incorporate visual material estimates for deformable linear object insertion~\cite{dloinsert}. Diffusion Policy further enables multimodal action generation for complex manipulation~\cite{diffusionpolicy}. However, sampled actions do not explicitly indicate their effect on subsequent alignment of the rod and the hole, which could cause damage of the rod or the hole. Visual predictive control mitigates the issue by evaluating candidate actions through the predicted radial and angular consequences before execution~\cite{visualforesight,predictiverep}.

\begin{figure}[t]
\centering
\includegraphics[width=\columnwidth]{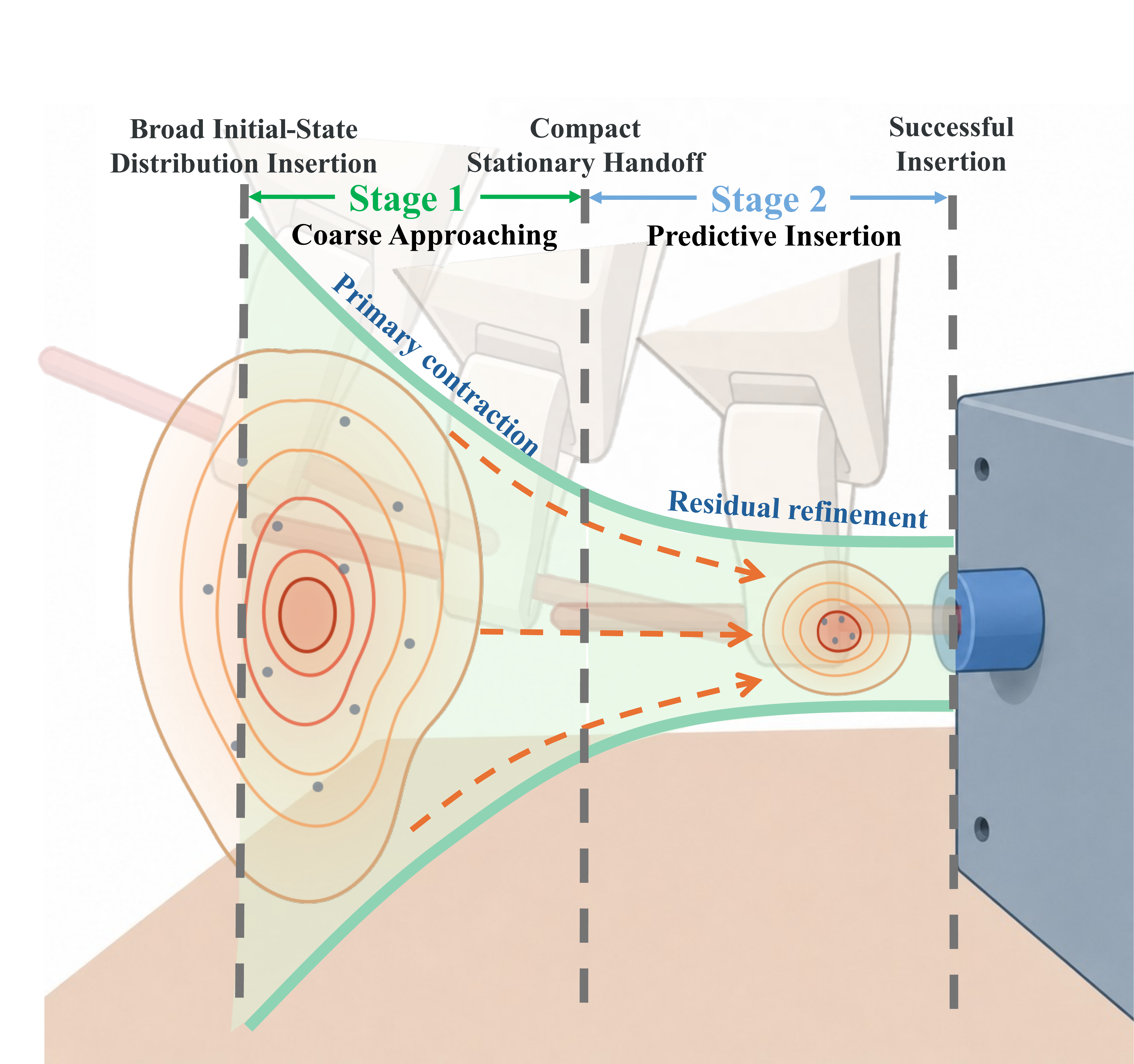}
\caption{Method overview and distribution shaping.  Crosses denote distal tip locations and the green envelope denotes state coverage.  Stage~1 contracts broad initial variation into a stationary hand-off; Stage~2 refines alignment while inserting.}
\label{fig:inference}
\end{figure}

While learning methods keep improving insertion success of traditional rigid peg-in-hole tasks, they cannot be readily transferred to slender rod insertion due to the perception and execution challenges induced by reduced volumes. On top of which, learning based methods are still constrained by data scarcity and poor out of distribution generalisation. These limitations are equally pertinent to slender rod insertion. To be specific, the initial pose of the robot and the rod could vary significantly, and regardless, they need to approach the hole for insertion. For a learning framework with limited demonstration data, it is impossible to enumerate all the possible initial poses. In contrast, the majority of insertion trajectories will share the near-hole area. This makes it straightforward to factorise the insertion tasks into two stages, i.e., one stage that guides the robot and the rod to a near hole hand-off area; and a second stage that carries out precise insertion. Fig.~\ref{fig:inference} shows the two-stage slender rod insertion strategy. It is worth noting that the two stages can be wrapped into an end-to-end setting.


We present \textit{RodForesight}, a world model enhanced diffusion framework for vision-guided insertion of slender rods across different materials. Visual servoing first guides the robot and rod to a compact near hole hand-off region. A diffusion policy then generates candidate action chunks for fine insertion, while an action-conditioned world model predicts their consequences in terms of tilt and radial alignment errors. The selected action is executed and the resulting state is reobserved, forming an iterative predict-execute-observe process. Both learned models only rely on visual observations without requiring explicit material labels.

The contributions of this work are threefold. First, we formulate millimetre-scale slender rod insertion as a vision-guided manipulation problem that explicitly accounts for deformation and material variation. Second, we introduce a two-stage framework that separates coarse approaching from predictive insertion, allowing \method{} to focus learning on fine near-hole corrections while using a world model to evaluate their expected effects before execution. Third, we conduct a range of experiments to evaluate both individual stages and the end-to-end setting, demonstrating improved insertion success across different rod materials and generalisation to unseen materials, including rods with stiffness outside the training range.

\section{Related Work}

\subsection{Insertion and Assembly with Contact}

In peg-in-hole assembly, contact provides both motion constraints and information about pose errors.  Kim and Rodriguez~\cite{activecontact} use tactile feedback to estimate contact and control subsequent interactions, while Chen et al.~\cite{robustpeg} use compliant motions to localise the hole and refine insertion.

For a flexible object, insertion also depends on how gripper motion changes the distal shape.  Geometric feedback and estimates of flexibility offer two ways to account for this dependence.  Wang et al.~\cite{tightrope} guide string and rope through narrow openings using an approximate Jacobian and a vector field, with regrasping to adjust the configuration.  Li and Choi~\cite{dloinsert} estimate flexibility from visual interaction cues and condition an insertion policy on that estimate.  Our task concerns slender rods, whose local alignment must be controlled across different material responses.

\subsection{Diffusion Policies for Robotic Manipulation}

Diffusion models represent a distribution over motions through iterative denoising.  Diffuser~\cite{diffuser} uses guided trajectory generation for planning, while Diffusion Policy~\cite{diffusionpolicy} generates action sequences conditioned on observations and repeatedly replans after partial execution.  This formulation also supports manipulation of flexible objects: Li et al.~\cite{dlorouting} combine reinforcement learning and a diffusion policy for routing deformable linear objects.  In \method{}, diffusion provides candidate corrections near the hole.  A separate world model predicts their effects on distal alignment, so the controller can compare proposals from the same observation before execution.

\subsection{World Models and Predictive Action Selection}

Predictive control requires a representation of how actions change the scene.  Visual foresight predicts future images for planning robot motions~\cite{visualforesight}, and predictive latent representations support planning for deformable objects~\cite{predictiverep}.  Object structure can also be represented explicitly: graph models propagate local interaction effects along deformable objects~\cite{dlodynamics}, and AdaptiGraph~\cite{adaptigraph} conditions graph dynamics on estimated material properties.  Differentiable rope and elastic rod models incorporate physical constraints into prediction and control~\cite{ropepbd,deform}.

For insertion, \method{} predicts the distal rod geometry relative to the hole and uses its tilt and radial errors to select an action prefix.  Training branches start from the same committed Cosserat state, so their targets describe the consequences of different candidate motions under identical initial conditions.  The world model therefore learns to predict the alignment quantities used by the controller.

\section{Problem Formulation}

The task involves inserting a flexible rod into a cylindrical hole using vision under initial pose variation, deformation, and contact.  The task is defined in a frame fixed to the hole whose positive $y$ axis points into it.  The distal rod geometry is represented by $\bg_t=[x_t,z_t,d_t,\theta_{x,t},\theta_{z,t}]^\top$, where $(x_t,z_t)$ is the transverse displacement, $d_t$ is insertion depth, and $(\theta_{x,t},\theta_{z,t})$ parameterises the tilt of the rod axis.  The radial and tilt errors are $r_t=\lVert[x_t,z_t]\rVert_2$ and $\phi_t=\lVert[\theta_{x,t},\theta_{z,t}]\rVert_2$, respectively.  This geometry defines task progress and supplies supervision and evaluation targets.

At control time $t$, the robot receives synchronised red–green–blue and depth (RGB-D) observations from side and wrist cameras.  Semantic masks for the rod, hole, hole holder, and gripper are extracted from these observations.  The action $\ba_t\in\mathbb{R}^6$ is a relative Cartesian gripper motion comprising three translations and three rotations.  The control objective is to advance the distal rod along the hole while reducing $r_t$ and $\phi_t$.  The controller uses camera observations to achieve the prescribed insertion depth while accommodating deformation caused by contact.

The initial approach and insertion near the hole present different control regimes.  During approach, large pose errors can be corrected from explicit geometry before contact.  Near the hole, smaller motions interact with rod deformation and contact, so visually plausible actions can produce different alignment outcomes.  Learning both regimes from complete expert rollouts also mixes many routine forward motions with comparatively few corrective motions.  This motivates a controller that first maps the broad initial set into a compact hand-off set and then makes repeated local decisions within that set.  Overall success is measured from the original initial condition and therefore includes Stage~1 failures.
\section{Methodology}

\subsection{Method Overview}

\method{} separates coarse approaching from predictive insertion, as shown in Fig.~\ref{fig:inference}. Stage~1 uses visual servoing to bring diverse initial poses into a compact region in front of the hole. After a stationary handoff, Stage~2 proposes local insertion motions with a diffusion policy and uses a learned world model to predict their effects on alignment. The controller executes a prefix of the selected sequence and then reobserves before replanning. The Cosserat model provides physical transitions for data generation and evaluation, while the learned world model supplies candidate predictions during control.

\subsection{Cosserat Rod Transition Model}
\label{sec:cosserat}

The simulator follows Cosserat rod kinematics~\cite{cosseratrod}, representing the rod by a centreline and an orientation at each cross section along $s\in[0,L]$, where $L$ is the rod length.  Nodal positions and rotations form the configuration $\mathbf q$.  For element $e$ of length $\ell_e$, the endpoint positions are $\mathbf r_e$ and $\mathbf r_{e+1}$. The midpoint orientation frame is $\mathbf C_e$, and $\boldsymbol\varphi_e$ is the logarithmic relative rotation vector between the end frames.  The element strain relative to its undeformed value $\boldsymbol\xi_{e,0}$ is
\begin{equation}
\boldsymbol\xi_e(\mathbf{q})=
\begin{bmatrix}
\mathbf{C}_e^\top
(\mathbf{r}_{e+1}-\mathbf{r}_e)/\ell_e\\
\boldsymbol\varphi_e/\ell_e
\end{bmatrix}
-\boldsymbol\xi_{e,0}.
\label{eq:cosserat_strain}
\end{equation}
The first vector block describes shear and extension, while the second describes bending and twist.  The material law relates these strains to internal forces and tangent stiffness through integration across the section. Axial and bending responses follow a bilinear elastoplastic law, while shear and torsion remain elastic.

Contact with the hole is computed from the surface gap, accounting for the rod radius. A logarithmic barrier resists penetration, and a Coulomb law governs tangential friction. The grasp constrains a band of the rod while leaving the remaining portion free to deform. Each transition is integrated with backward Euler and solved by Newton iterations with line search and adaptive step reduction. Once the solver converges, the resulting deformation and contact histories are committed for the next transition.

\subsection{Stage 1: Coarse Approaching}

Stage~1 estimates alignment from the segmented rod and hole. SAM (Segment Anything)~3~\cite{sam3} initialises the semantic masks, which XMem~\cite{xmem} then tracks across observations.  Calibrated depth is used to reconstruct the hole annulus and visible rod centreline.  The rod endpoint and axis provide its position and tilt relative to the hole.  These estimates are propagated using commanded tilt increments, then smoothed with separate position and tilt update rates.  Angular updates are bounded to limit abrupt changes in the estimate.

Bounded proportional corrections reduce lateral and angular errors while moving the rod towards a target depth $d_*$ in front of the hole.  Axial motion uses the attenuation factor
\begin{equation}
\alpha_t=\exp\!\left[-\left(\frac{\rho_t}{\sigma_r}\right)^2
-\left(\frac{\psi_t}{\sigma_\phi}\right)^2\right],
\label{eq:stage1_alignment}
\end{equation}
where $\rho_t$ and $\psi_t$ are the filtered radial and tilt errors, and $\sigma_r$ and $\sigma_\phi$ are their attenuation scales.  This reduces the approach speed when alignment is poor.  A tilt deadband suppresses small reversals.  Once radial, depth, and tilt errors meet the prescribed tolerances, the controller commands zero motion and waits for consecutive valid observations.  Stage~2 then starts from the resulting physical state and a fresh semantic history.  The tolerances and control settings are given in Sec.~\ref{sec:experimental_settings}.

\subsection{Stage 2: Predictive Insertion}

\begin{figure*}[t]
\centering
\includegraphics[width=\textwidth]{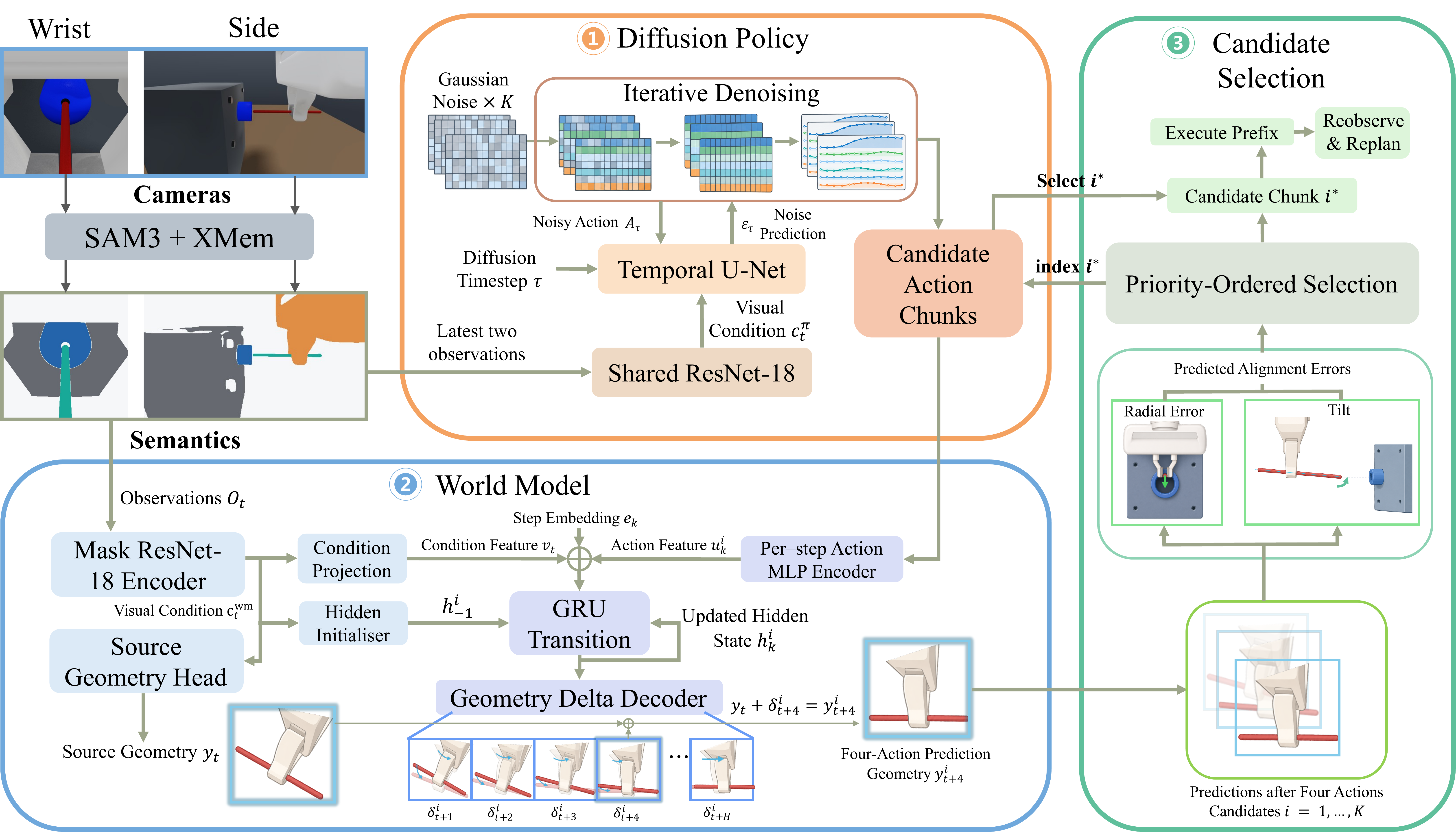}
\caption{Stage~2 inference. Paired semantic histories condition the diffusion policy and world model. The policy generates eight proposals of 16 actions. Selection uses the geometry predicted after four actions, giving priority to tilt and using radial error to resolve equal tilt values. The selected prefix is executed before reobservation and replanning. Insets illustrate the alignment errors.}
\label{fig:stage2_inference}
\end{figure*}

At each decision, Stage~2 samples $K$ candidate sequences of $H$ actions. Selection compares their predicted geometry after the first $H_e$ actions, matching the evaluation horizon to the prefix that will be executed. Fig.~\ref{fig:stage2_inference} shows the inference process.

The diffusion policy receives a history $\mathbf{o}_t$ of the $H_o$ most recent valid pairs of side and wrist semantic observations. This history is updated between executed prefixes. A ResNet-18 encoder~\cite{resnet}, shared across views and observation times, combines the masks with view and temporal embeddings to produce the visual condition $\bc_t^\pi$. A temporal U-Net~\cite{diffusionpolicy} uses this condition and the diffusion timestep to iteratively denoise action sequences from independent Gaussian samples. The policy learns these sequences from expert insertion trajectories. Decoding converts the normalised outputs to Cartesian translation and rotation increments, with axial motion restricted to the insertion direction.

The world model encodes the same semantic history with a separate mask encoder, initialised from the policy encoder and refined during training. From its visual condition $\bc_t^{\mathrm{wm}}$, the source geometry head estimates the current rod geometry relative to the hole, $\by_t$, using the coordinate order of $\bg_t$. The hidden state initialiser uses the same condition to initialise each candidate rollout.

Each candidate $i$ starts from a copy $\bh_{-1}^i$ of the same initial hidden state. The gated recurrent unit (GRU) transition $f_r$ then processes its actions in order, with step index $k$ starting from zero. Each update combines a projected visual condition $\mathbf v_t$, an encoded action $\bu_k^i$, and a learned step embedding $\mathbf e_k$:
\begin{equation}
\bh^{i}_{k}=f_r\!\left(
\bh^{i}_{k-1},\bu_k^i+\mathbf v_t+\mathbf{e}_k\right).
\label{eq:wm_transition}
\end{equation}

The geometry delta decoder $f_\delta$ maps each hidden state to the geometry change $\boldsymbol\delta^i_{t+k+1}$ over the full prefix processed so far. Adding this change to the shared source estimate gives the predicted geometry:
\begin{equation}
\boldsymbol\delta^{i}_{t+k+1}=f_\delta(\bh^{i}_k),\qquad
\by^{i}_{t+k+1}=\by_t+\boldsymbol\delta^{i}_{t+k+1}.
\label{eq:wm_geometry}
\end{equation}
During training, candidate branches start from the same expert state, including its deformation and contact history. Each branch executes a sequence sampled by the frozen diffusion policy, and the resulting geometry changes provide supervision at every prediction step. A single world model selected by validation loss is used at inference; training details are given in Sec.~\ref{sec:wm_training}.

At the execution horizon, the controller computes radial and tilt errors from each predicted geometry using the definitions in the problem formulation. It chooses the candidate with the smallest predicted tilt. Radial error resolves equal tilt values, followed by candidate index if both errors coincide. The policy supplies insertion motions learned from expert trajectories under the axial direction constraint, while selection prioritises angular alignment among those proposals.

The robot executes the first $H_e$ actions of the selected sequence, discards the remainder, and acquires a fresh observation for replanning. If an observation is invalid, it holds position and reacquires observations up to a fixed recovery budget. Insertion succeeds when the prescribed depth is reached with radial and tilt errors within tolerance. Solver nonconvergence or exhaustion of either the recovery or control budget ends the rollout in failure.

\section{Experiments}

Stage~1 uses bounded visual servoing.  In Stage~2, expert insertions supervise the policy, while its frozen samples and matched Cosserat rollouts supervise three world model replicas.  Validation loss selects one for inference.

\subsection{Experimental Instantiation}
\label{sec:experimental_settings}

Table~\ref{tab:setup} summarises the task geometry, evaluation range, and controller horizons.  Training and the principal evaluations span three materials, three grasp locations, and a $4\times2\times4$ grid of hole positions.

The Cosserat rod is discretised into 20 elements.  Contact stiffness $k_c$ is $10^4$~N/m, with static and dynamic friction coefficients of 0.20 and 0.15.

Stage~1 allows 100 motions.  Its position and tilt updates use gains of 0.65 and 0.35, respectively. Tilt is first propagated using the commanded rotation, and the tilt innovation vector is capped at a norm of $1^\circ$ before applying the update gain. The servo uses a gain of 0.65, with translational and angular command components in $[-1,1]$~mm and $[-1,1]^\circ$, respectively. Axial attenuation uses scales of 12~mm and $15^\circ$, while a tilt deadband of $0.5^\circ$ suppresses small reversals.

\begin{table}[bh]
\caption{Core task and evaluation settings.}
\label{tab:setup}
\centering
\scriptsize
\begin{tabularx}{\columnwidth}{@{}lY@{}}
\toprule
Setting & Value\\
\midrule
Rod and hole & Rod length 100~mm; rod diameter 4~mm; hole diameter 5~mm; hole depth 20~mm\\
Initial configuration & $x,z\in[-24,24]$~mm; $d\in[-85,-60]$~mm; $\theta_x,\theta_z\in[-20,20]^\circ$\\
Stage~1 hand-off & $d_*=-10$~mm; radial/depth/tilt tolerances $(1.5~\mathrm{mm},1.5~\mathrm{mm},3^\circ)$; five stationary frames\\
Stage~2 planning & Two semantic frames; eight proposals of 16 actions; four actions executed per decision\\
Successful insertion & Depth $\geq15$~mm, radial error $\leq0.5$~mm, and tilt $\leq2^\circ$\\
\bottomrule
\end{tabularx}
\end{table}

Stage~2 permits 160 controls at 50~ms intervals.  Transverse, axial, and angular action components lie in $[-0.5,0.5]$~mm, $[0,0.25]$~mm, and $[-1,1]^\circ$, respectively.

The policy uses an image condition of dimension 256 and a temporal U-Net with 256 channels.  Diffusion training uses 100 timesteps, whereas inference uses 20 reverse steps.  The world model uses an action embedding of dimension 128 and a GRU with 256 units.

\begin{table*}[b]
\caption{Success rates (\%) for controller ablations and generalisation tests. Stage~2 rates are conditional on successful Stage~1 handoffs; overall rates include all initial cases.  Dashes mark one stage controllers evaluated as a single process.}
\label{tab:candidate_ablation}
\centering
\scriptsize
\setlength{\tabcolsep}{2pt}
\begin{tabular}{@{}lllccc@{}}
\toprule
Structure & Action policy & World model & Stage~1(\%) & Stage~2(\%) & Overall(\%) \\
\midrule
\multicolumn{6}{@{}l}{\emph{Evaluation within the training distribution}} \\
One stage & Deterministic Transformer & No & -- & -- & 0.0 \\
One stage & Temporal U-Net diffusion & No & -- & -- & 1.1 \\
Two stages & Temporal U-Net diffusion & No & 100.0 & 88.9 & 88.9 \\
Two stages & Temporal U-Net diffusion & Transformer & 100.0 & 93.3 & 93.3 \\
Two stages & Temporal U-Net diffusion & GRU (\method{}) & 100.0 & \textbf{96.7} &
\textbf{96.7} \\
\midrule
\multicolumn{6}{@{}l}{\emph{Generalisation to hole positions and initial rod poses}} \\
Two stages & Temporal U-Net diffusion & GRU (\method{}) & 95.7 & 91.4 &
87.5 \\
\midrule
\multicolumn{6}{@{}l}{\emph{Generalisation to unseen rod diameters: trained at 2, 4, and 6~mm}} \\
Two stages, test: 3~mm & Temporal U-Net diffusion & GRU & 97.8 & 70.5 & 68.9 \\
Two stages, test: 5~mm & Temporal U-Net diffusion & GRU & 100.0 & 86.7 & 86.7 \\
\midrule
\multicolumn{6}{@{}l}{\emph{Generalisation to unseen materials: no retraining}} \\
Two stages, 6061-T6 aluminium & Temporal U-Net diffusion & GRU & 100.0 & 93.3 & 93.3 \\
Two stages, dry PA66 & Temporal U-Net diffusion & GRU & 100.0 & 93.3 & 93.3 \\
\bottomrule
\end{tabular}
\end{table*}

\subsection{Expert Data and Diffusion Policy Training}

A geometry expert generates 300 successful Stage~2 trajectories. Processing these trajectories with the same SAM~3 and XMem pipeline used during execution yields 6,654 training and evaluation samples. Each sample contains two semantic frames and a target sequence of 16 actions. The data are split by disjoint hole position cells, with 623 states held out. For sampling during training, phase labels group the examples into transition, insertion, and terminal strata.

Let $d_a$ denote the action dimension and $\mathbf{A}\in[-1,1]^{H\times d_a}$ an encoded expert action chunk.  The timestep $\tau$ is sampled uniformly from the diffusion times $\mathcal{T}$.  For Gaussian noise $\boldsymbol\epsilon\sim\mathcal{N}(\mathbf{0},\mathbf{I})$, let $c_\tau$ and $s_\tau$ denote the signal and noise coefficients obtained from the cumulative squared cosine schedule.  The noisy chunk is $\mathbf{A}_\tau=c_\tau\mathbf{A}+s_\tau\boldsymbol\epsilon$.  The policy learns to recover the sampled noise from this chunk and the visual condition:
\begin{equation}
\calL_\pi=\mathbb{E}\!\left[
\left\lVert
\boldsymbol\epsilon_\theta(\mathbf{A}_\tau,\bc_t^\pi,\tau)
-\boldsymbol\epsilon
\right\rVert_2^2\right].
\label{eq:dploss}
\end{equation}
Here, $\bc_t^\pi$ is the encoded policy condition with history length $H_o$.

The policy is trained for 80 epochs using AdamW with batch size 64, learning rate $10^{-4}$, and mixed precision.  Its weights are then frozen for branch generation.

\subsection{Matched Branches and World Model Training}
\label{sec:wm_training}

At each expert state, the frozen policy generates eight proposals of 16 actions. Each proposal is rolled out from the same observation and committed Cosserat state, producing 53,232 matched branches.  Because the source state is held fixed, differences in the outcomes reflect the candidate actions. These branches provide the world model with supervision for comparing candidate motions before execution.  Training branches originate from expert states, whereas the controller may visit different states during execution. Prediction accuracy is therefore also evaluated on states reached by the controller in Sec.~\ref{sec:wm_diagnostics}.

Each world model replica receives two mask frames and all eight action chunks, with source geometry and changes supervised at all 16 prediction steps.  Let $\mathcal K$ index the $H$ prediction steps, starting from zero; $\boldsymbol\delta_{\mathcal K}$ and $\Delta\bg_{\mathcal K}$ denote the predicted changes and their simulated targets, respectively. The coordinate scales $\mathbf{s}_g$ and $\mathbf{s}_\delta$ are $[2.5,2.5,15,10,10]$ and $[1,1,4,8,8]$, respectively, with position in millimetres and angle in degrees.  With coordinate division $\oslash$, the objective is
\begin{equation}
\begin{aligned}
\calL_m={}&\lambda_g\operatorname{SmoothL1}\!\left(
\by_t\oslash\mathbf{s}_g,\bg_t\oslash\mathbf{s}_g\right)\\
&+\lambda_\delta\operatorname{SmoothL1}\!\left(
\boldsymbol\delta_{\mathcal K}\oslash\mathbf{s}_\delta,
\Delta\bg_{\mathcal K}\oslash\mathbf{s}_\delta\right).
\end{aligned}
\label{eq:wmloss}
\end{equation}

The loss weights $\lambda_g$ and $\lambda_\delta$ are 0.5 and 1, respectively. Each replica is trained for 40 epochs with batch size 32 and an AdamW learning rate of $3\times10^{-4}$.  Sampling is balanced across phase and candidate diversity, and the mask encoder is refined from its diffusion initialisation. The replicas share architecture and hyperparameters but use independent initialisation and data order.  Only the replica with the lowest validation loss is retained for inference.

\begin{figure*}[bht]
\centering
\includegraphics[width=\textwidth]{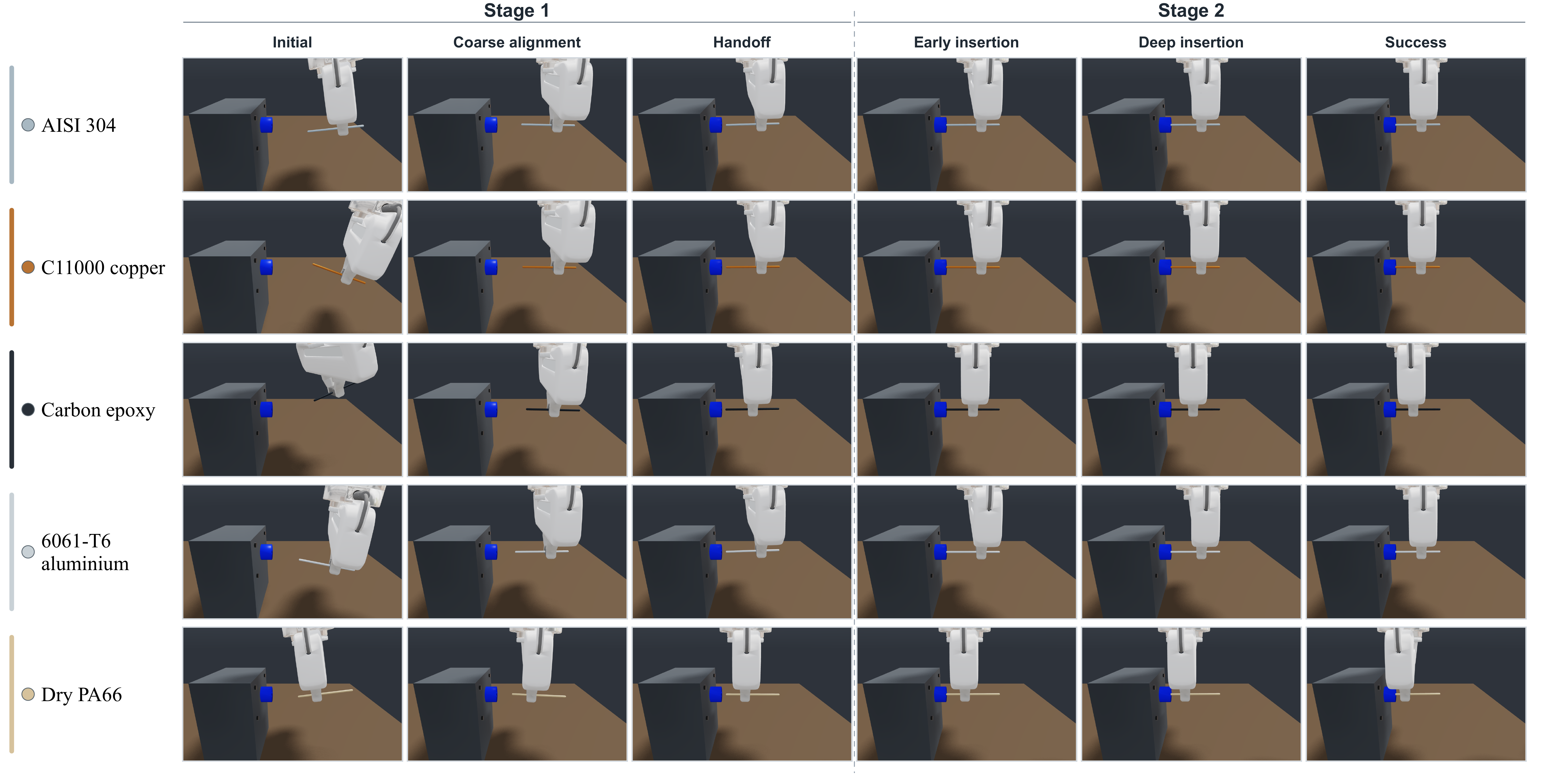}
\caption{Representative successful executions across five materials. The first three rows test unseen hole positions and initial rod poses, while the last two show 6061-T6 aluminium and dry PA66. Columns show the initial state, Stage~1 coarse approaching and handoff, Stage~2 insertion, and success. Colours identify materials for display.}
\label{fig:generalization_sequences}
\end{figure*}


\subsection{Evaluation Protocol}

Direct and staged controllers are compared on 90 fixed initial configurations within the training distribution. These comprise 29 AISI~304, 31 C11000 copper, and 30 carbon epoxy cases. For these paired comparisons, direct policies start from the initial poses, whereas staged variants restore the corresponding Stage~1 handoffs. Both direct baselines use 300 expert trajectories with phase sampling and validation selection over three runs. They generate sequences of 16 actions and execute four per decision. All variants acquire fresh observations and share physical states and diffusion seeds where applicable. Test cases are fixed before execution, and every failure remains in the overall denominator.

Generalisation is evaluated on 256 cases with unseen hole positions and a broader range of initial rod positions and orientations. The material counts are 88 for AISI~304, 87 for C11000 copper, and 81 for carbon epoxy. Each diameter test uses the same 90 initial configurations as the principal evaluation. Models train at 2, 4, and 6~mm and are evaluated at 3 and 5~mm, keeping radial clearance at 0.5~mm. The checkpoint is selected using the training validation set.

Material tests use 100~mm long, 4~mm diameter rods of 6061-T6 aluminium or dry PA66 and the same 90 initial configurations. Each rod is initialised straight and unstrained, with zero velocity and acceleration and no contact history, under zero external load. The Cosserat state evolves during Stage~1 and passes to Stage~2 with its configuration, motion, and internal and contact histories intact. Both learned models remain frozen, and the contact parameters, visual appearance, control settings, and sampling seeds remain unchanged. Overall success includes every initial case.

Code, model weights, and evaluation data are available in an anonymised repository.\footnote{\url{https://github.com/emotionalchara-lang/RodForesight-review}}

\subsection{Results}

\subsubsection{Stage Separation}

Separating coarse approaching from insertion improves success even before introducing a world model.  With the temporal U-Net diffusion policy, the staged controller achieves 88.9\% success, compared with 1.1\% when the policy controls the complete task directly (Table~\ref{tab:candidate_ablation}).  The deterministic Transformer trained by behaviour cloning achieves 0.0\% success.  Within this demonstration budget, the results support assigning broad pose correction to Stage~1 so that the policy can concentrate on local insertion motions.

\begin{table}[h]
\caption{Success rates (\%) across the three training materials. Every initial case remains in the denominator.}
\label{tab:material_breakdown}
\centering
\scriptsize
\setlength{\tabcolsep}{3pt}
\begin{tabular}{@{}lccc@{}}
\toprule
Configuration & AISI~304(\%) & C11000(\%) & Carbon (\%) \\
\midrule
\multicolumn{4}{@{}l}{\emph{Evaluation within the training distribution}} \\
One stage, Transformer policy & 0.0 & 0.0 & 0.0 \\
One stage, diffusion policy & 3.4 & 0.0 & 0.0 \\
Two stages, no world model & 93.1 & 93.5 & 80.0 \\
Two stages, Transformer model & 93.1 & 96.8 & 90.0 \\
Two stages, GRU model (\method{}) & \textbf{96.6} & \textbf{100.0} & \textbf{93.3} \\
\midrule
\multicolumn{4}{@{}l}{\emph{Generalisation to hole positions and initial rod poses}} \\
Two stages, GRU model & 86.4 & 89.7 & 86.4 \\
\bottomrule
\end{tabular}
\end{table}

\subsubsection{Predictive Action Selection}
\label{sec:wm_diagnostics}

The world model improves insertion by choosing among motions proposed by the same diffusion policy.  With the policy and initial cases fixed, predictive selection raises success from 88.9\% to 96.7\%, with all previously successful trials remaining successful.  The exact paired McNemar test gives a $p$ value of 0.0156, with gains across all three training materials (Table~\ref{tab:material_breakdown}). Replacing only the GRU transition with a Transformer gives 93.3\% success. However, the paired comparison with the GRU gives a $p$ value of 0.375, so these results do not establish superiority of either architecture.

\begin{table}[h]
\caption{World model prediction errors after four actions.
Radial errors are in millimetres and tilt errors are in degrees.
Generalisation refers to the 256 cases with unseen hole positions and initial rod poses.
Rows labelled `successful' include only successful rollouts.}
\label{tab:wm_prediction}
\centering
\scriptsize
\setlength{\tabcolsep}{2.5pt}
\begin{tabular}{@{}lcccc@{}}
\toprule
& \multicolumn{2}{c}{Radial (mm)} & \multicolumn{2}{c}{Tilt ($^\circ$)} \\
\cmidrule(lr){2-3}\cmidrule(lr){4-5}
State set or material & MAE & RMSE & MAE & RMSE \\
\midrule
\multicolumn{5}{@{}l}{\emph{Matched test branches}} \\
All materials & 0.028 & 0.052 & 0.108 & 0.225 \\
AISI~304 & 0.028 & 0.052 & 0.130 & 0.361 \\
C11000 & 0.029 & 0.054 & 0.101 & 0.155 \\
Carbon epoxy & 0.026 & 0.049 & 0.101 & 0.158 \\
\midrule
\multicolumn{5}{@{}l}{\emph{Executed branches from policy states}} \\
Training distribution, all & 0.576 & 3.208 & 0.567 & 0.822 \\
Training distribution, successful & 0.156 & 0.186 & 0.503 & 0.605 \\
Generalisation, all & 0.761 & 3.739 & 0.690 & 0.953 \\
Generalisation, successful & 0.163 & 0.193 & 0.568 & 0.669 \\
\bottomrule

\end{tabular}
\par\smallskip
\begin{minipage}{\columnwidth}
\scriptsize
\raggedright
\textit{Note:}
MAE and RMSE denote mean absolute error and root mean square
error, respectively.
\end{minipage}
\end{table}

\begin{table}[h]
\caption{Candidate selection on 623 test panels with matched branches.  The diverse subset contains 28 panels: 7 AISI~304, 14 C11000, and 7 carbon epoxy.  Lower regret is better.}
\label{tab:selection_diagnostics}
\centering
\scriptsize
\setlength{\tabcolsep}{2.2pt}
\begin{tabular}{@{}lcccc@{}}
\toprule
& \multicolumn{2}{c}{Exact match (\%)} & \multicolumn{2}{c}{Regret} \\
\cmidrule(lr){2-3}\cmidrule(lr){4-5}
Selector or material & All & Diverse & All & Diverse \\
\midrule
\multicolumn{5}{@{}l}{\emph{Selector using predicted geometry}} \\
All materials & 3.4 & 42.9 & 0.025 & 0.248 \\
AISI~304 & 4.5 & 57.1 & 0.017 & 0.088 \\
C11000 & 3.8 & 42.9 & 0.026 & 0.288 \\
Carbon epoxy & 1.3 & 28.6 & 0.029 & 0.328 \\
\midrule
No world model, all & 13.3 & 10.7 & 0.031 & 0.534 \\
\bottomrule
\end{tabular}
\end{table}

Matched candidate branches help identify when prediction improves action selection.  Exact index agreement counts differences even when candidates have nearly identical outcomes.  Regret instead measures the gap between the selected and oracle outcomes, normalised by the success tolerances. Prediction reduces mean regret from 0.031 to 0.025 across all test panels (Table~\ref{tab:selection_diagnostics}).  The reduction is larger on the 28 panels selected by a fixed action diversity threshold, where regret falls from 0.534 to 0.248.  This suggests that prediction is particularly useful when the candidate corrections have different consequences.

On expert branches, the world model has radial MAE in $[0.026,0.029]$~mm and tilt MAE in $[0.101,0.130]^\circ$ across the three training materials (Table~\ref{tab:wm_prediction}).  Accuracy decreases, however, on states reached by the controller.  Radial MAE rises to 0.576~mm in the training distribution test and 0.761~mm in the generalisation test. Together with the paired success comparison, these results show that prediction can improve action selection even though its accuracy decreases during execution.

\subsection{Ablation and Generalisation}

The frozen controller also generalises to softer materials. Stage~1 succeeds in every case for both 6061-T6 aluminium and dry PA66, and overall success is 93.3\% for each material (Table~\ref{tab:candidate_ablation}). All failures occur during Stage~2. Aluminium and PA66 have Young's moduli of 68.9 and 3.5~GPa, respectively, compared with the training range of $[117,193]$~GPa. With rod dimensions fixed, these results support transfer to lower stiffness without retraining, including the much softer PA66 model. The observed responses remain elastic, with PA66 represented by an isotropic surrogate. Fig.~\ref{fig:generalization_sequences} shows representative successful executions.

Stage~1 also reduces the larger initial errors in the test with unseen hole positions and initial rod poses.  Among successful handoffs, median radial error falls from 19.39 to 0.38~mm, while median tilt falls from $20.06^\circ$ to $1.20^\circ$.  Most of the initial misalignment is therefore corrected before the insertion policy takes over.  Stage~1 achieves 95.7\% success, while the complete controller achieves 87.5\%, with all initial cases retained.  These tests use the three training materials to assess generalisation in hole position and initial pose.

Transfer across rod diameters is less uniform.  After training at 2, 4, and 6~mm, the controller achieves 68.9\% success at 3~mm and 86.7\% at 5~mm, with radial clearance fixed.  The thinner rod remains difficult to observe: perception recovery accounts for 89.3\% of failures in the 3~mm test.  These results demonstrate insertion at diameters absent from training, while identifying perception as the principal bottleneck at 3~mm.

\section{Conclusion}

This manuscript presented \method{} for inserting long, thin flexible rods through fine clearances. Stage~1 contracts broad initial variation into a compact hand-off region, allowing the diffusion policy in Stage~2 to focus on contact-sensitive corrections. The world model evaluates candidate actions by predicting their radial and angular effects before execution. Experiments show that \method{} improves insertion success over baseline methods, including Diffusion Policy, and generalises across unseen hole positions, initial rod poses, rod diameters, and materials, including a highly flexible material with compliance well beyond the training range.

\balance
\bibliographystyle{IEEEtran}
\bibliography{references}

\end{document}